\pdfoutput=1
\documentclass[11pt]{article}

\usepackage[preprint]{acl}

\usepackage{color}

\usepackage{colortbl}
\definecolor{mygray}{gray}{0.8}

\usepackage{times}
\usepackage{latexsym}
\usepackage{enumitem}
\usepackage[normalem]{ulem}

\usepackage[T1]{fontenc}
\usepackage[utf8]{inputenc}

\usepackage{microtype}

\usepackage{inconsolata}

\usepackage{graphicx}

\usepackage{bbm}
\usepackage{multirow}
\usepackage{caption}
\usepackage{float}
\usepackage{subcaption}
\usepackage{colortbl}
\usepackage{xcolor}
\usepackage{balance}
\usepackage{multicol}
\usepackage{booktabs}
\usepackage{tabularx}
\usepackage{comment}

\title{\textsc{LLM-GAN}: Construct Generative Adversarial Network Through Large Language Models For Explainable Fake News Detection}

\author{
Yifeng Wang, Zhouhong Gu, Siwei Zhang, Suhang Zheng, Tao Wang, Tianyu Li, Hongwei Feng, Yanghua Xiao 
}

\author{Yifeng Wang\textsuperscript{\rm $\spadesuit$},
Zhouhong Gu\textsuperscript{\rm $\spadesuit$},
Siwei Zhang\textsuperscript{\rm $\spadesuit$},
Suhang Zheng\textsuperscript{\rm $\heartsuit$},
Tao Wang\textsuperscript{\rm $\heartsuit$},\\
\bf
Tianyu Li\textsuperscript{\rm $\heartsuit$},
Hongwei Feng\textsuperscript{\rm $\spadesuit$}\thanks{Corresponding authors.},
Yanghua Xiao\textsuperscript{\rm $\spadesuit$}\footnotemark[1]
\\
\textsuperscript{\rm $\spadesuit$}Shanghai Key Laboratory of Data Science, School of Computer Science, Fudan University\\
\textsuperscript{\rm $\heartsuit$}Alibaba Group\\
\texttt{\{yfwang20, hwfeng, shawyh\}@fudan.edu.cn},\\
\texttt{\{zhgu22, swzhang22\}@m.fudan.edu.cn}\\
\texttt{\{suhang.zhengsh, shayue.wt, qianchuan.lty\}@alibaba-inc.com}
}

\begin{document}
\newcommand{\model}{\textsc{{LLM-GAN}}}
\maketitle
\begin{abstract}
Explainable fake news detection predicts the authenticity of news items with annotated explanations. 
Today, Large Language Models (LLMs) are known for their powerful natural language understanding and explanation generation abilities. 
However, presenting LLMs for explainable fake news detection remains two main challenges. 
Firstly, fake news appears reasonable and could easily mislead LLMs, leaving them unable to understand the complex news-faking process. 
Secondly, utilizing LLMs for this task would generate both correct and incorrect explanations, which necessitates abundant labor in the loop. 
In this paper, we propose \textbf{{\model}}, a novel framework that utilizes prompting mechanisms to enable an LLM to become Generator and Detector and for realistic fake news generation and detection.
% {\model} consists of two main stages:
% (i) inter-adversary prompting, which offers semantic gradients for the Detector in an adversarial manner, enabling it to benefit from the news-faking process.
% (ii) self-reflection prompting that automates the Detector to revise itself from its past mistakes, thus eliminating the need for human annotators. 
Our results demonstrate {\model}'s effectiveness in both prediction performance and explanation quality. We further showcase the integration of {\model} to a cloud-native AI platform to provide better fake news detection service in the cloud.

% This deploys the Detector as an autonomous agent that can improve on its past mistakes through a self-reflective loop.

\end{abstract}

\section{Introduction}\label{sec:introduction}
The rapid spread of fake news on social media platforms has significant negative impacts on many real-world industries, such as politics \cite{politics}, economics \cite{ecnomics}, and public healthcare \cite{health}. To tackle this issue, fake news detection aims to predict the authenticity of news items, which has been widely developed in recent years \cite{fake1, ARG}.

Existing fake news detection methods \cite{emo, fakebert, fakeRL, fake1} predominantly utilize deep learning techniques to automatically extract and summarize news information. Despite their success, these methods still face two primary limitations. Firstly, existing methods struggle to fully understand and distinguish fake news due to the intricate news-faking process, where the news-fakers can manipulate any aspect of the news. They cannot analyze the hidden motivations behind the fake news, thus leading to suboptimal predictions for effectively capturing such a stochastic news-faking process.
Another issue is their failure to address the explainability of predictions. This reduces their applicability in real-world industries as users could doubt the prediction results \cite{explainable}. Explainable fake news detection allows users to understand the reasoning process of systems, enabling users to make their own inferences and thereby boosting their trust \cite{trust}.

Today, Large Language Models (LLMs) \cite{LLMsurvey_1, LLMservey, gu2024xiezhi} offer a promising solution to these problems for their powerful natural language understanding and generation capabilities in both text classification \cite{textclassification} and explanation generation \cite{LLMforexplain}. 
Currently, leveraging LLMs for explainable fake news detection remains unexplored. \textit{Our work aims to fill this gap by addressing the challenges of prompting an LLM to achieve explainable fake news prediction:}

\begin{figure}[t]
    \centering
    \includegraphics[width=\linewidth]{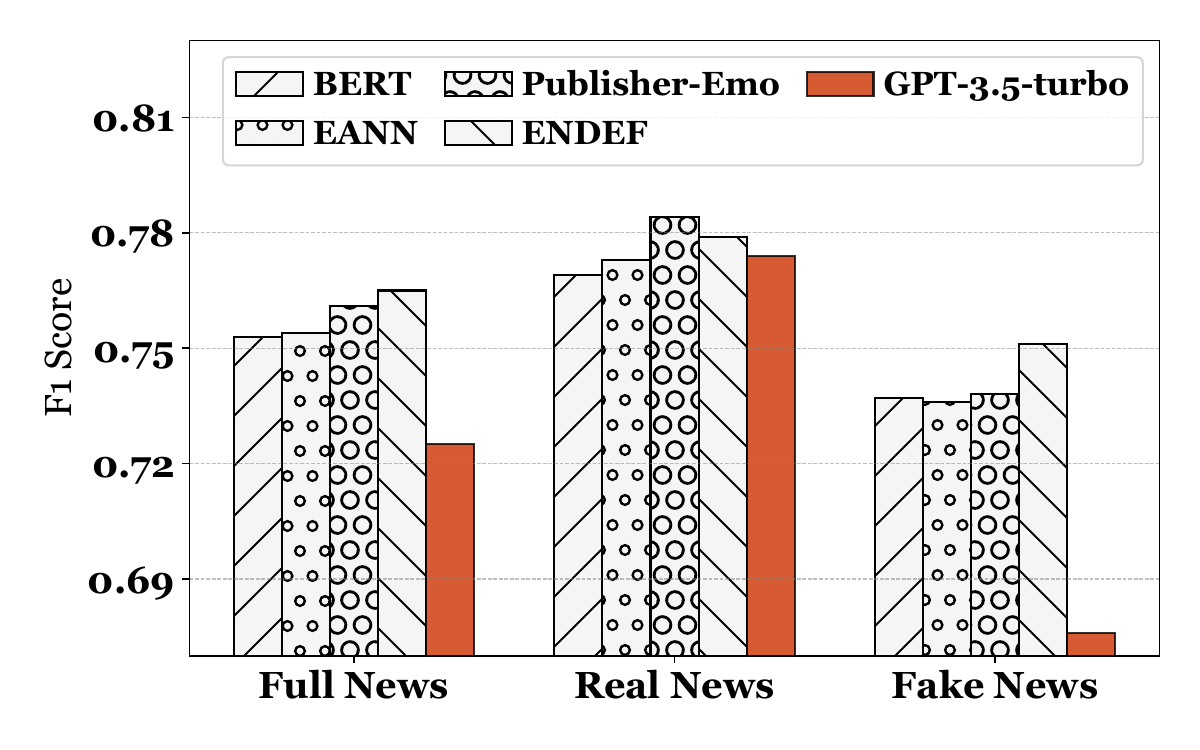}
    % %\vspace{-8mm}
    \caption{Comparisons of performance in fake news detection on the Weibo21 dataset. The simply prompted LLM (orange) underperforms compared to existing deep-learning-based methods, especially when predicting fake news. For more details, please refer to Sec. \ref{sec:experiments}.
    }
    \label{fig:introduction_2}
    %\vspace{-5mm}
\end{figure}

Firstly, fake news often appears reasonable and coherent \cite{fake2}, and can easily deceive and mislead LLMs. This makes it challenging for LLMs to detect potential logical errors or misinformation in news items, leading to confirmatory biases \cite{LLMbias} where they tend to classify news as real. 
We conduct preliminary experiments to test how prompted LLMs perform for fake news detection in Fig. \ref{fig:introduction_2}. We find that simply prompted LLMs underperform compared to existing deep-learning-based methods, with an even larger performance gap when predicting fake news. 
Therefore, it requires LLMs to understand and adapt to the news-faking process, a capability they currently lack.
Secondly, the problem becomes increasingly complex with the introduction of explainability requirements, as it demands LLMs to verbally justify why a news item is fake or not. However, when prompting LLMs for this task, they can generate both correct and incorrect explanations for each news item. This necessitates a substantial amount of labor from journalism experts, making it both costly and impractical.

To address the aforementioned issues, we propose \textbf{{\model}}.
{\model} utilizes prompting mechanisms to enable an LLM Detector to realize explainable fake news detection, which compromises two main stages:
(i) inter-adversary prompting.
Inspired by adversarial networks \cite{gan}, we introduce an LLM Generator to generate highly deceptive fake news, fooling the Detector and thus compelling it to learn from the news-faking process. This mechanism prompts both the Generator and the Detector to continually refine their strategies, offering semantic gradients and promoting their capabilities in an adversarial interplay.
(ii) self-reflection prompting.
To automate the detection process, we propose self-reflection prompting that teaches the Detector via a verbal self-reflective manner. Specifically, we allow the Detector to obtain its own past mistakes with evidenced explanations, which can be used as prompting samples and automatically revise itself without any human intervention. 

To investigate the model's effectiveness, we conduct extensive experiments across real-world datasets for explainable fake news detection. 
{\model} outperforms existing fake news detection methods and annotates clear explanations for its predictions. In summary, our main contributions are as follows:
% To investigate the effectiveness of {\model}, we conduct extensive experiments across real-world datasets. Our model outperforms existing fake news detection methods with comprehensive explanations for its detections. 

%\vspace{-4mm}
\begin{itemize}[itemsep=0pt, parsep=0pt, leftmargin=*]
    \item \textbf{Idea:} We focus on investigating the challenges of LLMs for explainable fake news detection. To the best of our knowledge, it is the first attempt to prompt LLMs for the explainable fake news detection task successfully.
    \item \textbf{Solution:} We propose {\model} model, a solution that utilizes inter-adversary and self-reflection prompting mechanisms to enable an LLM Detector to learn from both news-faking process and its past mistakes in a fully autonomous manner.
    \item \textbf{Products:} In addition to validating {\model}’s effectiveness in both prediction performance and explanation quality, we showcase the integration of {\model} to an industrial product to provide better explainable fake news detection service.
\end{itemize}

%\vspace{-5mm}

\begin{figure*}[t]
    \centering
    \includegraphics[width=0.94\linewidth]{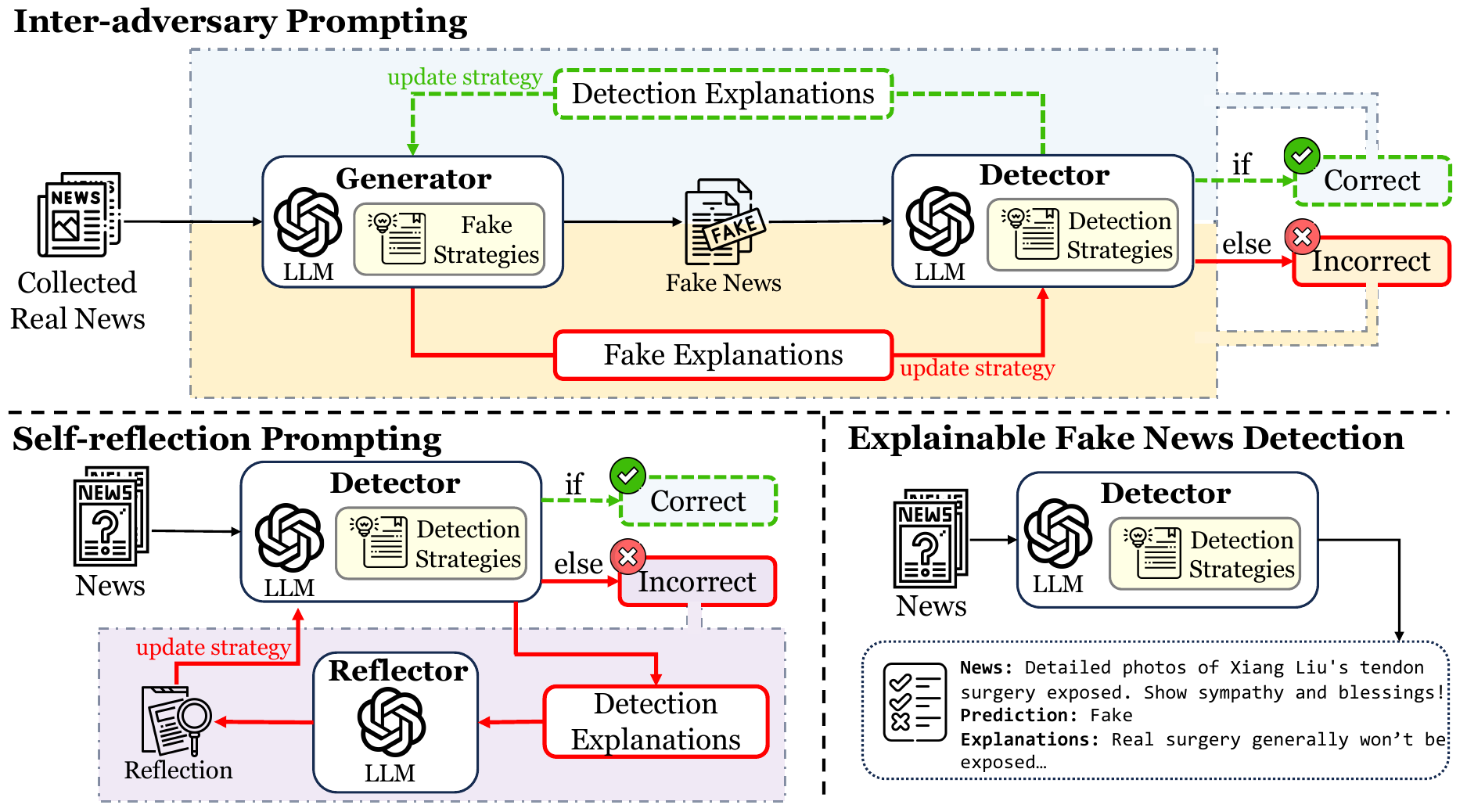}
    \caption{
    The architecture of our proposed {\model} model for explainable fake news detection. {\model} consists of two main stages: (i) inter-adversary prompting that allows the Detector to benefit from the news-faking process; and (ii) self-reflection prompting that automates the Detector to revise itself from its past mistakes.}
    \label{fig:overall}
    %\vspace{-5mm}
\end{figure*}

%\vspace{-5mm}
\section{Related Work}\label{sec:relatedwork}
%\vspace{-3mm}
\textbf{Fake News Detection.}
Most existing methods for fake news detection rely on deep learning techniques, and they can broadly categorized into two types: social-based and individual-based methods. Social-based methods focus on leveraging the social context of news, such as propagation patterns \cite{pattern}, user reactions \cite{user}, and social networks \cite{social}. In contrast, individual-based methods directly analyze the news content, including text \cite{text}, images \cite{image}, or additional knowledge \cite{bases, enviroment}. However, these methods overlook explainability and cannot provide robust evidence to support their predictions. In this paper, we utilize LLMs for explainable fake news detection, offering accurate predictions with clear explanations. 

\noindent\textbf{Language Models for Data Generation and Augmentation.}
Data generation and augmentation leverages generative language models \cite{generatedmodel} to create expressive data and enhance various natural language understanding tasks such as QA \cite{QA} and NIL \cite{NIL}. Some methods concentrate on data generation \cite{generation_1, generation_2, generation_3, generation_4}, while others explore augmentation \cite{augmentation_1, augmentation_2, augmentation_3, gu2023gantee}.
Different from these existing methods based on general tasks (e.g., QA), we focus on the task-specific generation and utilize LLMs to generate fake news and evidenced explanations, augmenting the LLM Detector for explainable fake news detection.

\section{Methodology}
%\vspace{-3mm}
In this section, we first introduce the problem formulation and data augmentation in this paper. As illustrated in Fig. \ref{fig:overall}, we then present our {\model} model for explainable fake news detection.  

%{\model} prompts an LLM Detector via two main steps: (i) inter-adversary prompting, which enables the Detector to learn from the news-faking process, and (ii) self-reflection prompting that allows the Detector to automatically teach and revise itself in a self-reflective manner.

%\vspace{-3mm}
\subsection{Preliminaries}\label{sec:preliminaries}
%\vspace{-1mm}

\textbf{Problem Formulation.}
Given a news item $x \in \mathcal{X} = \left\{x_i\right\}_{i=1}^O$, explainable fake news detection aims to predict $\hat{\mathcal{Y}}_x = (\hat{y}_x, \hat{\mathbf{e}}_x)$, where $\hat{y}_x = \left\{0, 1\right\}$ represents the authenticity prediction (whether the news item is fake or not), and $\hat{\mathbf{e}}_x$ is the explanation. 

\noindent\textbf{News Collection and Augmentation.}
Recall that in the inter-adversary prompting mechanism, our Generator creates fake news and fools the Detector into learning from the news-faking process. 
To enhance the quality of fake news generation, we conduct data augmentation by adopting the data collection techniques used for the Weibo21 dataset \cite{weibo21}, a popular benchmark in many fake news detection methods \cite{EANN, emo, ARG}. 
Specifically, We collect 6,739 real news items in the Sina News website\footnote{\url{https://news.sina.com.cn/}} from January to April 2024. These 
collected news items have a maximum token count of 639, a minimum of 27, and an average of 266 tokens.
% \footnote{We will make the data publicly available if this paper can get accepted.}.
We feed them into our Generator and guide it to generate highly misleading fake news, thus augmenting the whole news-faking process.

%\vspace{-2mm}
\subsection{Inter-adversary Prompting}
%\vspace{-1mm}
Inter-adversary prompting mechanism utilizes fake news to prompt a Detector $\mathrm{M}_D$ to summarize and refine its detection strategies $\mathcal{S}_D$. To achieve this, we introduce an LLM Generator $\mathrm{M}_G$ that exploits its fake strategies $\mathcal{S}_G$ to generate highly misleading fake news and fool the Detector via an adversarial process, compelling Detector to upgrade its detection strategies and enhance detection abilities. 

\noindent\textbf{Strategy Prompting.} 
The Generator creates fake news with the fake strategies, and we prompt it through two inputs: a specific collected real news item $x$ mentioned in Sec. \ref{sec:preliminaries}; and the fake strategies $\mathcal{S}_G$ obtained from the previous inter-adversary process. The Generator $\mathrm{M}_G$ produces fake news item $x'$ with provided fake explanations $\hat{\mathbf{e}}'_{x}$ by:
%\vspace{-2mm}
\begin{equation}
%\vspace{-2mm}
    x', \hat{\mathbf{e}}'_{x} = \mathrm{M}_G \left(x, \mathcal{S}_G\right).
\end{equation}
We then prompt the Detector $\mathrm{M}_D$ with the generated fake news and enable it to respond with the predictions and detection explanations, promoting its ability to effectively detect fake news as:
%\vspace{-2mm}
\begin{equation}
%\vspace{-2mm}
    \hat{y}_{x'}, \hat{\mathbf{e}}_{x'} = \mathrm{M}_D\left(x', \mathcal{S}_D\right),
\end{equation}
where $\hat{y}_{x'}=\left\{0, 1\right\}$ represents whether news item $x'$ is fake or not, and $\hat{\mathbf{e}}_{x'}$ the detection explanations. Notice that we follow ReAct \cite{react} and employ similar prompt trajectories, which incorporate a unified and integrated prediction-explanation step within the strategy prompting mechanism. 

\noindent\textbf{Inter-adversary Process.}
Inspired by adversarial networks \cite{gan}, we construct an adversarial architecture to allow the Detector to learn from the Generator's news-faking process. It establishes a strategic adversarial balance between the Generator and the Detector, enabling them to automatically refine and upgrade their strategies for generating and detecting fake news, respectively.

From the predictions made by the Detector $\hat{y}_{x'}$, we can obtain binary feedback based on the ground truth, i.e., correct or incorrect. When the Detector correctly detects the news, we consider that the Detector's capabilities surpass that of the Generator in generating fake news. Therefore, we enhance the Generator $\mathrm{M}_G$ and prompt it using Detector's detection explanations $\hat{\mathbf{e}}_{x'}$. As a result, the Generator can upgrade its fake strategies and guide it to produce more sophisticated fake news as follows:
%\vspace{-2mm}
\begin{equation}
    \mathcal{S}_G = \mathrm{M}_G\left(x, \hat{\mathbf{e}}_{x'}\right). %\vspace{-2mm}
\end{equation}
Conversely, if the Detector $\mathrm{M}_D$ fails to correctly detect the news, this suggests that the Detector's detection abilities are relatively poor, and we prompt it with the fake explanations $\hat{\mathbf{e}}'_{x}$ from the Generator, refining detection strategies and thereby learning from the news-faking process. We formalize it as:
%\vspace{-2mm}
\begin{equation}
%\vspace{-2mm}
    \mathcal{S}_D = \mathrm{M}_D\left(x, \hat{\mathbf{e}}'_{x}\right).
\end{equation}
These upgraded strategies, $\mathcal{S}_G$ for the Generator and $\mathcal{S}_D$ for the Detector, are obtained and then utilized to respectively generate fake news and make predictions in the next terms of inter-adversary prompting mechanism.

%\vspace{-2mm}
\subsection{Self-reflection Prompting}\label{sec:self-reflectionpromting}
%\vspace{-1mm}
Since LLMs are not specifically trained for fake news detection, there is a risk that the Detector may produce incorrect examples. To address this issue, we propose self-reflection prompting to allow the Detector to automatically revise itself from its past incorrect examples. To be specific, we prompt the Detector $\mathrm{M}_D$ by using its detection strategies $\mathcal{S}_D$ to predict a given news item $x$ in the training data of the datasets in Sec. \ref{sec:datasetandbaseline}, generating the predictions and detection explanations as follows:
%\vspace{-2mm}
\begin{equation}
%\vspace{-2mm}
    \hat{y}_{x}, \hat{\mathbf{e}}_{x} = \mathrm{M}_D\left(x, \mathcal{S}_D\right).
\end{equation}
It is crucial to note that the detection strategies are upgraded in the latest inter-adversary prompting process. To realize self-reflection prompting, once incorrect samples occur, we introduce an LLM Reflector $\mathrm{M}_R$ and prompt it with incorrect samples to generate reflection feedback $r_x$. This feedback should clearly explain where the Detector went wrong in its detection explanations and provide evidence that supports the authenticity of the news item. The overall formalization is:
%\vspace{-2mm}
\begin{equation}
%\vspace{-3mm}
    r_x = \mathrm{M}_R\left(x, \hat{\mathbf{e}}_{x}\right).
\end{equation}
The reflection feedback then feeds into Detector $\mathrm{M}_D$ for self-reflection to promptly refine its detection strategies, enabling the model to automatically revise itself from its past mistakes as follows:
%\vspace{-2mm}
\begin{equation}
%\vspace{-3mm}
    \mathcal{S}_D = \mathrm{M}_D\left(x, r_x\right).
\end{equation}
Consequently, we can obtain a Detector $\mathrm{M}_D$ with well-prompted detection strategies $\mathcal{S}_D$ for effective explainable fake news detection.

\begin{table*}[t]
    \centering
    \small
    \caption{Prediction performance comparisons for fake news detection between our {\model} model and the baselines. The best results are highlighted in \textbf{bold}, and the second-best are marked with \underline{underline}.}
    \label{tab:mainresults}
    \begin{tabular}{clcccccccc}
        \toprule
        \multicolumn{2}{c}{\multirow{3}{*}{\textbf{Methods}}} & \multicolumn{4}{c}{\textbf{Weibo21}} & \multicolumn{4}{c}{\textbf{GossipCop}} \\
        \cmidrule(lr){3-6} \cmidrule(lr){7-10}
        \multicolumn{2}{c}{} & macF1 & Acc. & F1-real & F1-fake & macF1 & Acc. & F1-real & F1-fake \\
        \midrule
        \multirow{6}{*}{\shortstack{Deep-Learning-based \\ Methods}} & BERT & 0.753 & 0.754 & 0.769 & 0.737 & 0.765 & 0.862 & 0.916 & 0.615 \\
        & EANN & 0.754 & 0.756 & 0.773 & 0.736 & 0.763 & 0.864 & 0.918 & 0.608 \\
        & Publisher-Emo & 0.761 & 0.763 & 0.784 & 0.738 & 0.766 & 0.868 & 0.920 & 0.611 \\
        & ENDEF & 0.765 & 0.766 & 0.779 & 0.751 & 0.768 & 0.865 & 0.918 & 0.618 \\
        & SuperICL & 0.757 & 0.759 & 0.779 & 0.734 & 0.736 & 0.864 & 0.920 & 0.551 \\
        & ARG & \underline{0.784} & \underline{0.786} & \underline{0.804} & \underline{0.764} & \underline{0.790} & \underline{0.878} & \underline{0.926} & \underline{0.653} \\
        \midrule
        \multirow{2}{*}{LLMs} & GPT-3.5-turbo & 0.725 & 0.734 & 0.774 & 0.676 & 0.702 & 0.813 & 0.884 & 0.519 \\
        & {\model} & \textbf{0.804} & \textbf{0.806} & \textbf{0.812} & \textbf{0.796} & \textbf{0.823} & \textbf{0.896} & \textbf{0.934} & \textbf{0.712}  \\
        \bottomrule
    \end{tabular}
    %\vspace{-4mm}
\end{table*}

%\vspace{-2mm}
\section{Experiments}\label{sec:experiments}
%\vspace{-2mm}
\subsection{Datasets and Baselines}\label{sec:datasetandbaseline}
%\vspace{-1mm}
\textbf{Datasets.}
We select two widely used datasets for fake news detection, including the Chinese dataset Weibo21 \cite{weibo21} and the English dataset GossipCop \cite{fake2}. We follow ARG \cite{ARG} and split the datasets for training, validating, and testing, respectively. Due to page limitations, detailed information of datasets is presented in the Sec. \ref{app:datasets} of the Appendix.

\noindent\textbf{Baselines.}
Most existing fake news detection methods rely on deep learning, and we select six representative methods, including BERT \cite{bert}, EANN \cite{EANN}, Publisher-Emo \cite{emo}, ENDEF \cite{ENDEF}, SuperICL \cite{superICL}, and ARG \cite{ARG}. Additionally, we prompt GPT-3.5-turbo \cite{gpt} via CoT \cite{COT} with few-shot demonstrations for fake news detection. Details are shown in Sec. \ref{app:baselines} of the Appendix.

\definecolor{lightgray}{gray}{0.8}
\begin{figure*}
    \centering
    \includegraphics[width=0.90\linewidth]{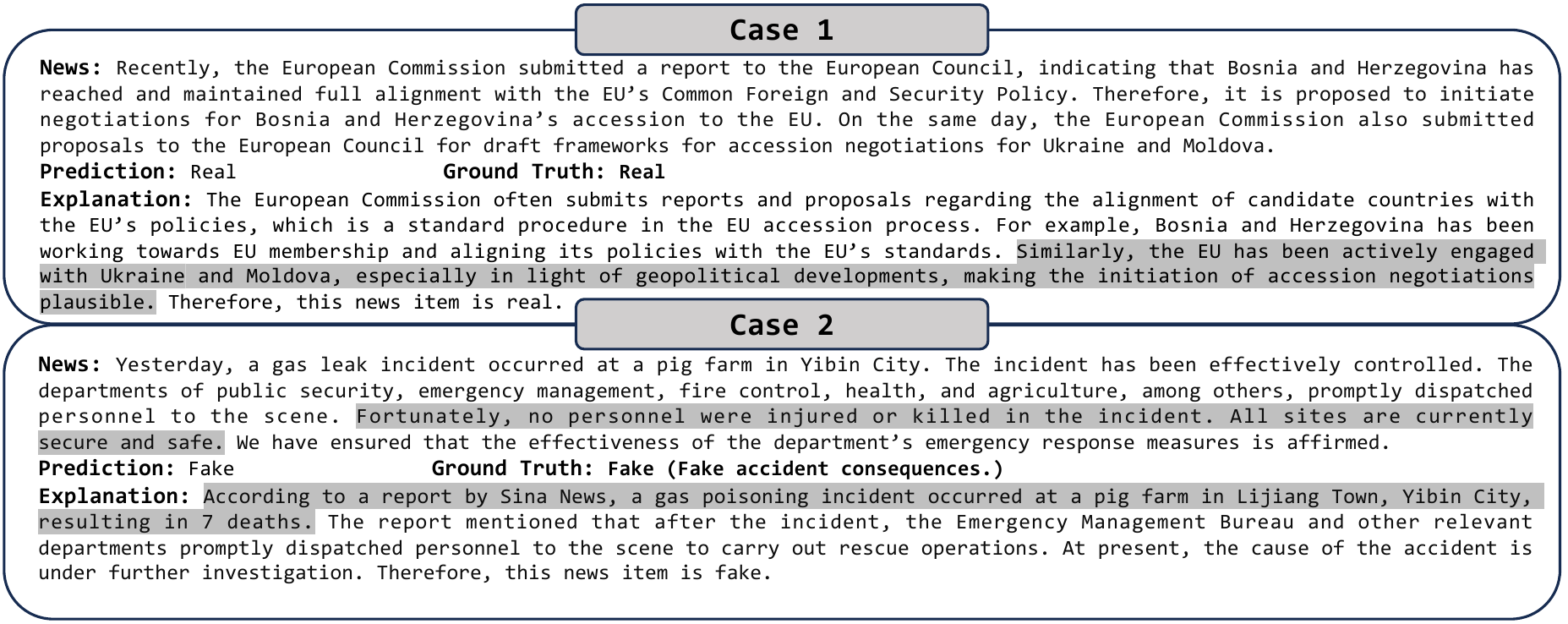} 
    \caption{Illustrative output examples of {\model} for two cases where the strongest baseline model, ARG \cite{ARG}, failed. {\model} makes correct predictions and provides logical, human-readable explanations as evidence. Errors in fake news and key information in explanations are highlighted with a \colorbox{lightgray}{gray background}.}
    \label{fig:case}
    %\vspace{-5mm}
\end{figure*}

%\vspace{-2mm}
\subsection{Explainable Fake News Detection}\label{sec:fakenewsdetection}
%\vspace{-1mm}
In this section, we first compare the prediction performance between {\model} and baselines, and then evaluate the explanation quality of the {\model}.

\noindent\textbf{Prediction Performance.}
Tab. \ref{tab:mainresults} reports the prediction performance for fake news detection. From the results, we find that:
(i) {\model} outperforms all baselines across all datasets and evaluation metrics, demonstrating its effectiveness.
(ii) GPT-3.5-turbo underperforms compared to existing deep-learning-based methods. A possible reason is that the simply prompted LLMs may not become an effective detector to replace task-specific deep-learning-based methods in fake news detection. In contrast, $\model$ successfully prompts LLMs to achieve superior performance, further validating its effectiveness.
(iii) Compared to GPT-3.5-turbo, the largest performance improvements of {\model} are observed when identifying fake news, which indicates that {\model} can successfully learn from the news-faking process and automatically revise itself from mistakes.

\noindent\textbf{Explanation Quality.}
In addition to the outstanding prediction performance, {\model} leverages the strengths of LLMs to provide explanations and achieves explainable fake news detection. Here, we analyze the explanation quality of our {\model}.

We can observe two main benefits from the provided explanations. Firstly, the explanations enable users to understand the {\model}'s decision-making process. Specifically, we select two cases where the strongest baseline model, i.e., ARG, failed, and present the corresponding predictions and explanations of {\model} in Fig. \ref{fig:case}. We find that {\model} can effectively analyze the content of the news and provide correctly logical, human-readable detection explanations, demonstrating its explainability.

The second benefit comes from the higher quality of explanations. To quantitively evaluate explanation quality, we propose a set of quality metrics and use GPT-4o to score each explanation from 1 to 7. We emphasize that such comparisons are fair since they evolve the same inputs. The average scores are shown in Tab. \ref{tab:score}. Interestingly, {\model} is not trained or prompted explicitly on these metrics, but it outperforms GPT-3.5-turbo across all these metrics. It indicates that our {\model} can help LLMs holistically improve their consideration of these factors, achieving better explanation quality.

%\vspace{-3mm}
\subsection{Ablation Study}\label{sec:ablationstudy}
%\vspace{-1mm}
In this section, we separately investigate the effectiveness of {\model}'s component, i.e., inter-adversary prompting and self-reflection prompting.

\begin{table}
    \centering
    \small
    \caption{Comparisons of explanation quality between our {\model} model and GPT-3.5-turbo. The scores are given by GPT-4o based on relevant metrics.}
    \label{tab:score}
    \resizebox{\columnwidth}{!}{
    \begin{tabular}{lcc}
        \toprule
        \textbf{Metrics} & \textbf{GPT-3.5-turbo} & \textbf{{\model}} \\
        \midrule
        Relevance to Detection & 4.1 & \textbf{5.7} \\
        Fairness of Real \& fake & 4.5 & \textbf{5.5} \\
        Accuracy for Detection & 4.7 & \textbf{6.1} \\
        Fact checking & 3.8 & \textbf{5.2} \\
        Integrity & 4.0 & \textbf{5.3} \\
        Contextual Understanding & 5.8 & \textbf{6.0} \\
        Clarity \& Coherence & 5.7 & \textbf{5.8} \\
        Consistency of Information & 5.3 & \textbf{5.9} \\
        Sensitivity to Updates & 4.3 & \textbf{5.5} \\
        \bottomrule
    \end{tabular}}
    %\vspace{-5mm}
\end{table}

\noindent\textbf{Inter-adversary Prompting.}
The inter-adversary prompting performs fake news generation and augmentation that utilizes our collected real news to generate misleading fake ones. Thus, we conduct an ablation study by controlling the number of input collected real news items for inter-adversary prompting mentioned in Sec. \ref{sec:preliminaries}, and report the macF1, F1-real, and F1-fake scores, respectively. When the input news number is zero, it means that we remove the inter-adversary prompting.

\begin{figure}
\centering
    \includegraphics[width=0.85\linewidth]{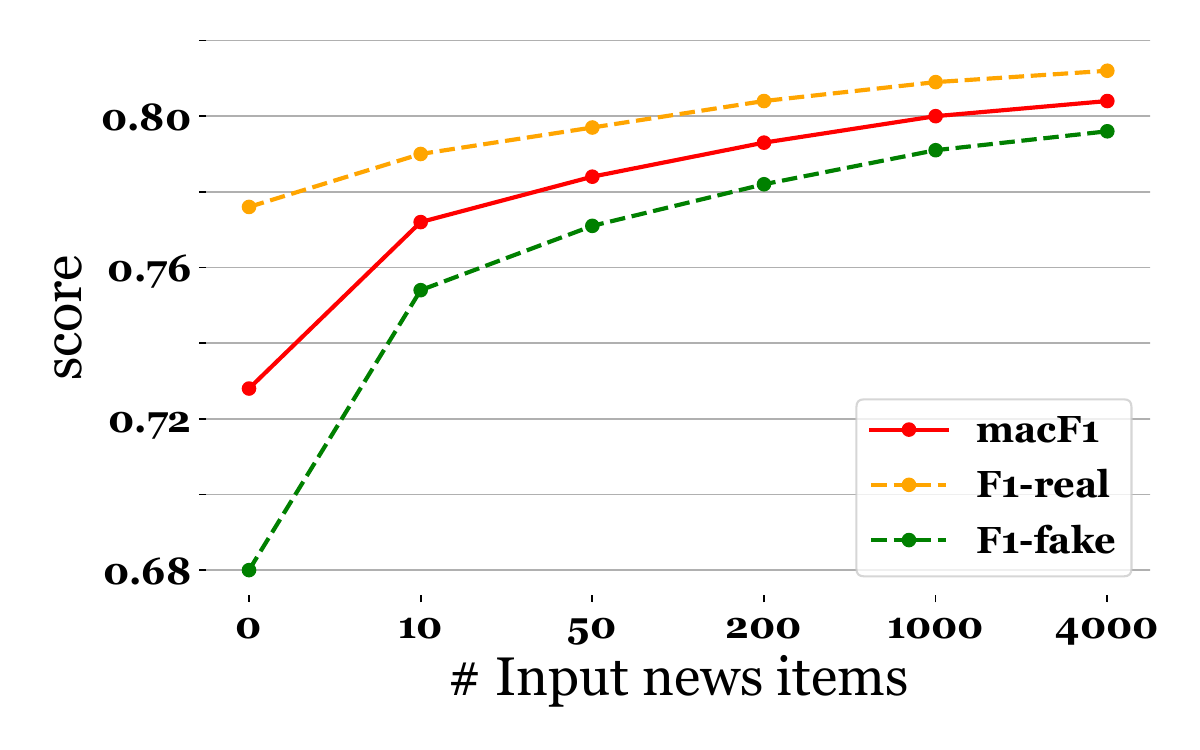}
    %\vspace{-5mm}
    \caption{Ablation study for inter-adversary prompting mechanism with the varying number of input collected news items. {\model} can effectively benefit from the news-faking process via data augmentation of the input news.}
    \label{fig:ablation}
    %\vspace{-5mm}
\end{figure}

The results for the Weibo21 dataset are presented in Fig. \ref{fig:ablation}. We see that with more inputs of real news for inter-adversary prompting, the model makes more and more correct predictions and demonstrates increasing performance. This evidences the effectiveness and exceptional robustness of the inter-adversary prompting module. Moreover, from the F1-fake score, the performance in identifying fake news is extremely boosted, highlighting the importance of adapting to the news-faking process. 

\noindent\textbf{Self-reflection Prompting.}
We conduct an ablation study with different prompting techniques used for the Reflector as shown in Tab.~\ref{tab:promptingtechniques}.
Zero-shot prompting constructs prompts with the task description and the given news, while few-shot prompting additionally includes news labels. CoT  (Chain-of-Though)~\cite{COT} adds eliciting sentences at the end of the prompt.
Besides, we remove the self-reflection prompting component, resulting in the variant ``w/o. SR''.

From the results in Tab. \ref{tab:promptingtechniques}, we find that directly removing the self-reflection prompting will result in significant performance degradation, demonstrating its effectiveness and necessity. Moreover, the few-shot variants outperform the zero-shot versions, indicating that using labels as prompt samples could enhance detection capabilities. Furthermore, CoT prompting generally provides certain performance improvements, particularly in the few-shot setting on the GossipCop dataset. However, we also observe instances where CoT leads to reduced performance, which suggests that effectively leveraging CoT may require more careful design.

\begin{comment}
\begin{figure}
  \centering
  \begin{subfigure}{0.8\linewidth}
    \centering
    \includegraphics[width=\linewidth]{figures/ablation.pdf}
  \end{subfigure}
  \centering
  \begin{subfigure}{0.8\linewidth}
    \centering
    \includegraphics[width=\linewidth]{figures/ablation.pdf}
  \end{subfigure}
  \caption{Ablation.}
  \label{fig:para}
\end{figure}
\end{comment}

\begin{table}
    \centering
    \small
    \caption{Ablation study for self-reflection (SR) prompting mechanism with various prompting techniques.}
    \label{tab:promptingtechniques}
    \begin{tabular}{>{\centering\arraybackslash}p{2.2cm}>{\centering\arraybackslash}p{0.85cm}>{\centering\arraybackslash}p{0.85cm}>{\centering\arraybackslash}p{0.85cm}>{\centering\arraybackslash}p{0.85cm}}
        \toprule
        \multirow{2}{*}{\shortstack{\textbf{Prompting} \\ \textbf{techniques}}}& \multicolumn{2}{c}{\textbf{Weibo21}} & \multicolumn{2}{c}{\textbf{GossipCop}} \\
        \cmidrule{2-5}
        & macF1 & Acc. & macF1 & Acc. \\
        \midrule
        zero-shot & 0.785 & 0.787 & 0.799 & 0.876 \\
        few-shot & \underline{0.798} & \underline{0.799} & \underline{0.809} & \underline{0.881} \\
        zero-shot CoT & 0.783 & 0.785 & 0.795 & 0.871 \\
        few-shot CoT & \textbf{0.804} & \textbf{0.806} & \textbf{0.823} & \textbf{0.896} \\
        w/o. SR & 0.760 & 0.761 & 0.765 & 0.852  \\
        \bottomrule
    \end{tabular}
    %\vspace{-5mm}
\end{table}

%\vspace{-2mm}
\section{Industrial Application}\label{sec:industry}
%\vspace{-2mm}
In this section, we briefly discuss how our {\model} benefits users in real-world industries. For a better user experience, we have integrated {\model} into a cloud-native AI platform (Platform of Artificial Intelligence, Alibaba Cloud\footnote{\url{https://www.alibabacloud.com/zh/product/machine-learning}}), which enables users (especially journalists and rumor experts) to detect fake news using various fake news detection methods. Users can input news content through the WebUI and freely choose the method they are interested in for detection. They can also view their previously submitted news and detection results. As illustrated in Fig. \ref{fig:industry}, users who select our $\model$ receive not only predictions, like those provided by traditional methods, but also additional human-readable explanations, significantly boosting their trust in the system. Moreover, based on the Queries Per Second demand and system workload, our inference service can automatically scale to a configurable number of computers on a GPU cluster, ensuring efficient and scalable performance.

%\vspace{-2mm}
\section{Conclusion}\label{Conclusion}
%\vspace{-2mm}
In this paper, we focus on prompting LLMs for explainable fake news detection and propose the {\model} model. {\model} not only enables an LLM Detector to benefit from the news-faking process through inter-adversary prompting, but also allows it to effectively revise itself from its past mistakes via a self-reflective process. Extensive experiments demonstrate {\model}'s effectiveness in both prediction performance and explanation quality. To boost user experience, we integrate {\model} into a fake news detection system, increasing user trust significantly.

% In this paper, we focus on prompting LLMs for explainable fake news detection, which is a significantly difficult task for existing deep-learning-based detection methods. We highlight two key challenges: the limitation of current LLMs in understanding the news-faking process; and their inability to automatically learn from past mistakes to provide explanations. To address these challenges, we propose the {\model}, a general LLM prompting paradigm for explainable fake news detection. This framework not only enables an LLM Detector to benefit from the news-faking process through inter-adversary prompting, but also allows it to effectively revise itself by learning from its own past mistakes via a self-reflective process. Extensive experimental results demonstrate the effectiveness of {\model} in both prediction performance and the quality of detection explanations. 

% \clearpage
% \newpage
% Bibliography entries for the entire Anthology, followed by custom entries
%\bibliography{anthology,custom}
% Custom bibliography entries only
\bibliography{custom}

% \clearpage
% \newpage

\appendix
\section{Appendix}
The codes and the datasets in this paper will be available if the paper is accepted.
% XXX\footnote{url will be given after this paper gets accepted.}

%\vspace{-2mm}
\subsection{Detailed Description of Datasets}\label{app:datasets}
We utilize two datasets for evaluation: Weibo21 and GossipCop. To void potential performance inflation due to data leakage, we follow existing deep-learning-based methods \cite{ARG, EANN, emo} and split the datasets for training, validating, and testing, respectively. The detailed statistics for datasets are summarized in Tab. \ref{tab:datasets}. For a fair comparison, we only utilize the training data for few-shot prompting when performing the {\model} model. 

\begin{table}[h]
    \centering
    \caption{Statistics of split distribution for two datasets.}
    \label{tab:datasets}
    \resizebox{\columnwidth}{!}{
    \begin{tabular}{>{\centering\arraybackslash}p{0.6cm}>{\centering\arraybackslash}p{0.7cm}>{\centering\arraybackslash}p{0.7cm}>{\centering\arraybackslash}p{0.7cm}>{\centering\arraybackslash}p{0.7cm}>{\centering\arraybackslash}p{0.7cm}>{\centering\arraybackslash}p{0.7cm}}
        \toprule
        \multirow{2}{*}{\textbf{\#}}& \multicolumn{3}{c}{\textbf{Weibo21}} & \multicolumn{3}{c}{\textbf{GossipCop}} \\
        \cmidrule{2-7}
        & Train & Val & Test & Train & Val & Test \\
        \midrule
        Real & 2,331 & 1,172 & 1,137 & 2,878 & 1,030 & 1,024 \\
        Fake & 2,873 & 779 & 814 & 1,006 & 244 & 234 \\
        Total & 5,204 & 1,951 & 1,951 & 3,884 & 1,274 & 1,258 \\
        \bottomrule
    \end{tabular}
    }
    %\vspace{-5mm}
\end{table}

\subsection{Detailed Description of Baselines}\label{app:baselines}
In this paper, we select six deep-learning-based methods as our baselines for comparison. We describe their detailed information as follows:
\begin{itemize}[itemsep=0pt, parsep=0pt, leftmargin=*]
    \item \textbf{BERT} \cite{bert} is a widely used pre-trained deep-learning-based model in natural language understanding, and we fine-tune it for the fake news detection task.
    \item \textbf{EANN} \cite{EANN} learns from the auxiliary signals. It is designed to minimize the impact of event-related features. For the auxiliary task, we use the publication month as the label.
    \item \textbf{Publisher-Emo} \cite{emo} analyzes the emotions behind the news comments and integrates them with textual features to enhance fake news detection.
    \item \textbf{ENDEF} \cite{ENDEF} aims to eliminate entity bias through causal learning, enhancing generalization on distribution-shifted fake news data.
    \item \textbf{SuperICL} \cite{superICL} enhances language models by using deep-learning-based plug-in models. For each test sample, it injects both the prediction and confidence level into the prompt.
    \item \textbf{ARG} \cite{ARG} advances deep-learning-based detection by providing adaptive rationale guidance, achieving state-of-the-art performance in fake news detection.
\end{itemize}
\begin{figure}
    \centering
    \includegraphics[width=\linewidth]{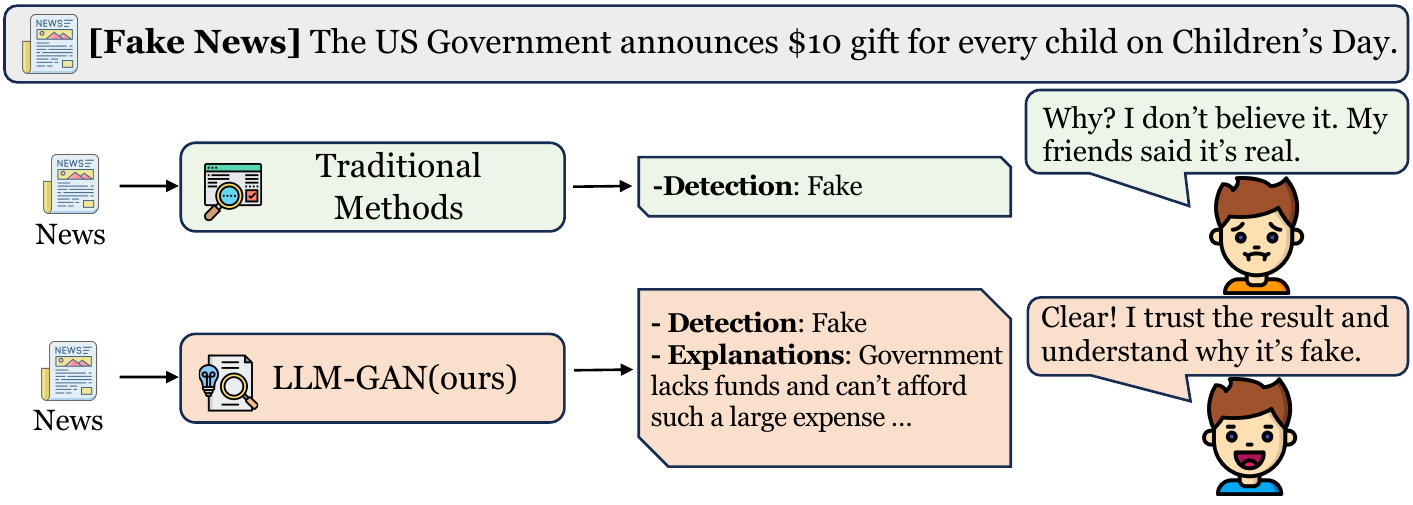}
    \caption{Comparisons of industrial application between {\model} and baselines. Our {\model} can significantly boost user trust in the system.}
    \label{fig:industry}
    %%\vspace{-5mm}
\end{figure}

In addition to the above deep-learning-based methods, we also prompt \textbf{GPT-3.5-turbo} \cite{gpt} for fake news detection as an LLM baseline. It is an LLM developed by OpenAI that powers the popular chatbot ChatGPT. We choose it for its representativeness and ease of use.

\begin{figure*}
    \centering
    \includegraphics[width=0.90\linewidth]{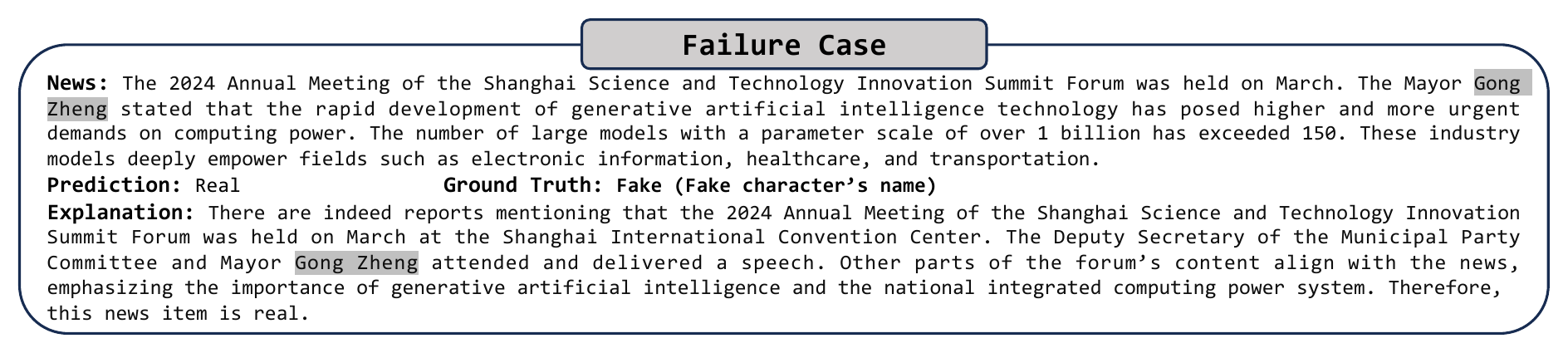} 
    \caption{Failure case made by the {\model}. Extremely hidden fake news may lead to incorrect prediction using {\model}. Errors in fake news and key information in explanations are highlighted with a \colorbox{lightgray}{gray background}.}
    \label{fig:case_app}
    %\vspace{-5mm}
\end{figure*}

\subsection{Detailed Description of Implementations}\label{app:detals}
We describe the detailed implementations used in our paper as follows:
\begin{itemize}[leftmargin=*, itemsep=0pt]
    \item \noindent\textbf{Baselines.}
For the deep-learning-based baselines, we follow ARG \cite{ARG} and limit the maximum text length to 300 tokens, and we conduct \textit{chinese-bert-wwm-ext} for Chinese and \textit{bert-base-uncased} for English evaluations from the Transformers package \cite{package}. We employ Adam as the optimizer and conduct a grid search to find the optimal learning rate, reporting the test results from the best-validation checkpoint. For GPT-3.5-turbo, to achieve fake news detection, we follow CoT \cite{COT} and prompt it with few-shot demonstrations to enable task learning.
\item \noindent\textbf{Ours.}
For our {\model}, we employ the GPT-3.5-turbo APIs for the Generator, Detector, and Reflector. All strategies within these LLM agents are initially set to null. The collected real news items in this paper are used as the inputs of the inter-adversary prompting, and the training news items of the public datasets are used to prompt LLMs with few-shot demonstrations in the self-reflection prompting.
\item \noindent\textbf{Evaluation Metrics.}
For comparisons, we utilize macro F1 score (macF1) and accuracy (Acc.) as our evaluation metrics. Furthermore, to explore the models' effectiveness in predicting real and fake news separately, we present the F1 score for the two classes (F1-real/F1-fake), respectively. 
\end{itemize}

\subsection{Limitation}
We present two limitations of this work as follows:
\begin{itemize}[leftmargin=*]
    \item This work focuses on text-based news. However, in real-world applications, many news items are multimodal \cite{multimodel_1} where they can contain images, audio, or video from their contents. We have not considered these modalities due to the simplicity of text content, and it is straightforward to integrate multimodal information into the model to enhance detection performance.
    \item  Another limitation of our work is the incorrect predictions of complex fake news.
    Fig. \ref{fig:case_app} demonstrates a failure case that {\model} made. It could be owing to the extremely subtle news-faking process of the given news: they forged the news by changing the character's name, and {\model} may struggle to detect such minor differences, thus leading to misclassification. Therefore, it is essential to research and design models with stronger detection capabilities.
\end{itemize}

\subsection{Ethics Statement}
In this study, we utilize datasets from public sources without the involvement of any human annotators. The rights associated with the used dataset remain the sole property of the original rights holders. This study is intended solely for academic research purposes.

Using LLMs for fake news detection has potential ethical and societal implications. Here, we highlight some of these issues and propose possible solutions for mitigating them when deploying our model for practical use in the industries
\begin{itemize}[leftmargin=*, itemsep=0pt]
    \item \textbf{Manipulation Risks.} Public opinion manipulation has always been a problem in news dissemination. Considering the known vulnerabilities of LLMs, such as jailbreaking prompts \cite{jai}, using LLMs for fake news generation might strengthen this risk. To mitigate these issues, measures should be implemented to thoroughly check user inputs before processing them through the model. Access to the LLM's internal knowledge base should be restricted to authorized users only.
    \item \textbf{Misinformation.} While the goal of explainable predictions is to generate trustworthy results, LLMs can also be used to produce deceptive misinformation \cite{disinformation}. Measures should be taken to verify the accuracy of facts in explanations, whether through automated verification or human review before they are presented as part of the explanation.
\end{itemize}
% \cite{factcheck}
In summary, the most effective mitigation strategy is to incorporate humans into the loop to address various potential problems. While LLMs can assist in performing labor-intensive tasks, they cannot replace the need for human oversight. Human involvement is crucial to ensure the accuracy, reliability, and ethical use of LLM-generated content.

%\vspace{-1mm}
\subsection{Discussion}
Our findings in explainable fake news detection reveal the limitations of LLMs in complex, real-world applications. Despite their impressive natural language understanding and analysis capabilities, LLMs struggle to fully leverage their parametric knowledge for fake news detection. This suggests that exploiting their potential may require more sophisticated prompting methods and a deeper understanding of their internal mechanisms. Based on this, we identify two major challenges for current LLMs in explainable fake news detection and provide a tailored solution: a prompting paradigm that enables LLMs to automatically induce and summarize their detection strategies, thus making accurate predictions and providing clear explanations. We hope our solution can be extended to other tasks and foster more efficient and cost-effective use of LLMs in the future.

\end{document}